\documentclass[12pt]{article}
\usepackage{amsmath}
\usepackage{graphicx,psfrag,epsf}
\usepackage{enumerate}
\usepackage{algorithm,algpseudocode}
\usepackage{float} 
\usepackage{url} 
\usepackage{cite}
\usepackage{array} 
\usepackage{amssymb}

\newcommand{\blind}{0}

\begin{document}

\def\spacingset#1{\renewcommand{\baselinestretch}%
{#1}\small\normalsize} \spacingset{1}


\if0\blind
{
  \title{\bf Robust Particle Swarm Optimizer based on Chemomimicry}
  \author{Casey Kneale, Karl S. Booksh\hspace{.2cm}\\
    Department of Chemistry and Biochemistry\\
    University of Delaware, Newark, Delaware 19716\\
	}
  \maketitle
} \fi

\if1\blind
{
  \bigskip
  \bigskip
  \bigskip
  \begin{center}
    {\LARGE\bf }
\end{center}
  \medskip
} \fi


\noindent%

\spacingset{1.45} 

Particle swarm optimizers (PSO) were first introduced by Kennedy and Eberhart as stochastic algorithms which seek optimal solutions to functions through the use of swarm intelligence \cite{eberhart:05}. The main theme of PSO is that many particles are allowed to \textit{explore} a function space. As each particle relocates it inputs its coordinates into the objective function for evaluation. Standard PSO methods assign particle directions and magnitudes for movement based on distances to the best global functional evaluation outcome ($\overrightarrow{g}$), and/or their individual best locations ($\overrightarrow{p}$). Traditionally the positions ($\overrightarrow{X}$) of particles and their velocities ($\overrightarrow{V}$) are updated as follows,
\begin{equation}
	\overrightarrow{X}_{i} = \overrightarrow{X}_{i}  + \overrightarrow{V}_{i}
\end{equation}
\begin{equation}
\overrightarrow{V}_{i} = \overrightarrow{V}_{i}  + c_1 \cdot runif(0,1) \cdot (\overrightarrow{p}_i - \overrightarrow{X}_i) + c_2 \cdot runif(0,1) \cdot (\overrightarrow{g}_i - \overrightarrow{X}_i)
\end{equation}
Where $i$ denotes an individual particle, $c$ represents user-defined weighting factors, and $runif(0,1)$ generates a uniformly distributed random number between 0 and 1 \cite{Mauricio:13}. Despite the fact that PSO is a widely used heuristic optimizer, standard particle swarm algorithms are not especially robust to local minima in multimodal functions \cite{Esquivel:13}. More-so, standard particle swarm methods can require very large numbers of functional iterations. For costly function evaluations this practice is undesirable.

This work introduces a simple, robust, and efficient PSO algorithm which is loosely based on the physical process of crystallization. The Crystallization Particle Swarm Optimizer (CPSO), originally developed for surface plasmon resonance sensor optimizations has similar update equations as the standard method (Equations 1 and 2). The main deviations from the traditional movement scheme that were implemented in the CPSO algorithm were that local communications were removed ($c_1 = 0$) and that the L$_1$ displacement factor was replaced by a scaling vector ($\overrightarrow{e}_i$) in the velocity update equation. 
\begin{equation}
\overrightarrow{V}_{i} =  \pm c_2  runif(0,1) \cdot \overrightarrow{e}_i
\end{equation}
The vector $e$ can be used to relate the experimentally attainable resolutions of the independent variables to the step sizes explored by the optimizer, or a reasonable value given the problem space. While the value and sign of $c_2$ was assigned based on whether or not the particles current position was greater than the known global best and the phase of motion the particles were undergoing. 

The three phases of motion used in the CPSO method were: diffusion, directed motion, and nucleation. The diffusion phase randomly positions particles from their previous locations, without information which pertains to a global best location, so that initial placements of the particles can be explored within the function space. During the directed motion phase the balance between random motion and motion towards the global best gradually trades until nucleation/convergence toward a global optima inevitably occurs. The iteration count for where these phases begin and end are user defined. However, these phases can be manually tuned by observing a dynamic scalar term, the chaos factor, that influences the random motion of particles which follows the complementary error function (Figure 1). This chaos factor, is implemented by multiplication with the velocity of each particle.

\begin{figure}[H]
	\begin{center}
		\includegraphics[width=3in]{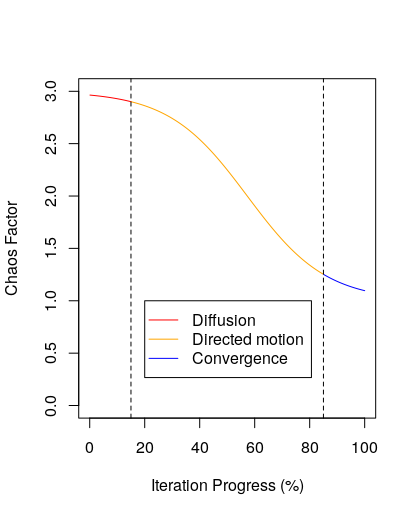}
	\end{center}
	\caption{Chaos factor (in units of $e$) vs iteration progress. \label{fig:first}}
\end{figure}

It should be stated that optimized spatial partitioning methods such as orthogonal nearest neighbor repulsive agent optimization \cite{Kneale:17}, or centroidal voronoi tessellations, should be employed in order for the diffusion phase to efficiently span the initial locations of the particles. This practice has been shown to improve the results of high dimensional PSO test functions \cite{Rich:04}, but we found that it improves the efficacy of the CPSO method. One final influencing factor was introduced to this algorithm which allowed for more informed searches during the nucleation stage. When particles were within a preset distance ($\leq 2 \overrightarrow{e}_i$) of the global best their velocity contribution towards the best was set to zero so that only random motion remained. This allowed for a robust, but random, search at the presumed global optima. The pseudo code for the algorithm is included in the appendix of this document.


\section{Results and Discussion}
\label{sec:data}
The CPSO algorithm was compared with the standard PSO2011 algorithm \cite{Mauricio:13} introduced by Clerc et al on the following test functions in ${\rm I\!R}^3$: griewank, rosenbrock, rastrigin, and parabola. Both algorithms attempted to find the global minima in multimodal and unimodal test functions with 1000 functional evaluations for each particle (30 replicates). The population sizes were varied from 5 to 20 particles in increments of 5 and the resulting optimized values were recorded (Table I).

Every multimodal function which was optimized by the CPSO algorithm featured lower means and standard deviations of the objectives than the respective trials performed with the PSO2011 algorithm. However, the PSO2011 algorithm performed considerably better on the unimodal parabola function. The order of magnitude discrepancies between CPSO and PSO2011 on the parabola test funcation may be explained by the scaling parameters employed. It is possible that with tuning of the CPSO parameters  better optimized values could result. However, for this report the values were left untuned to ensure a fair comparison. Despite the fact that CPSO was limited to the parameters that were explored, this method often returned mean solutions which were orders of magnitude lower than that of a standard algorithm with fewer particles. 

The efficacy of CPSO appeared to be linked to the manner in which it handles multimodal function spaces. The diffusion step is hypothesized to be the most important condition in overcoming convergence toward local minima. Future theoretical investigations and empirical tests on this algorithm will be reported elsewhere.

\begin{table}[H]
	\begin{center}
		\caption{Mean and standard deviations of results obtained from test functions } \label{tab:tabone}
		\begin{tabular}{r|ccccc}
Function	&\textbf{Griewank}	&	&	&	&\\
&Population	&5	&10	&15	&20\\
CPSO	&mean	&\textbf{1.0E-04}	&\textbf{4.7E-05}	&\textbf{4.5E-05}	&\textbf{2.4E-05}\\
&sd	&\textbf{6.8E-05}	&\textbf{3.5E-05}	&\textbf{3.0E-05}	&\textbf{1.7E-05}\\
PSO2011	&mean	&0.30	&0.15	&0.16	&0.16\\
&sd	&0.21	&8.9E-02	&6.4E-02	&5.0E-02\\
Function	&\textbf{Rosenbrock}	&	&	&	&\\
&Population	&5	&10	&15	&20\\
CPSO	&mean	&\textbf{0.97}	&\textbf{0.47}	&\textbf{0.42}	&\textbf{0.13}\\
&sd	&\textbf{0.81}	&\textbf{0.67}	&\textbf{0.52}	&\textbf{0.37}\\
PSO2011	&mean	&3.1	&1.2	&0.85	&0.57\\
&sd	&2.7	&1.7	&1.5	&1.3\\
Function	&\textbf{Rastrigin}	&	&	&	&\\
&Population	&5	&10	&15	&20\\
CPSO	&mean	&\textbf{7.7E-02}	&\textbf{4.1E-02}	&\textbf{2.3E-02}	&\textbf{1.6E-02}\\
&sd	&\textbf{3.9E-02}	&\textbf{3.2E-02}	&\textbf{2.1E-02}	&\textbf{1.2E-02}\\
PSO2011	&mean	&3.7	&1.9	&1.8	&1.6\\
&sd	&3.0	&1.1	&2.0	&1.1\\
Function	&\textbf{Parabola}	&	&	&	&\\
&Population	&5	&10	&15	&20\\
CPSO	&mean	&3.8E-04	&2.2E-04	&1.2E-04	&6.9E-05\\
&sd	&1.8E-04	&1.7E-04	&1.1E-04	&6.8E-05\\
PSO2011	&mean	&\textbf{1.4E-06}	&\textbf{6.5E-09}	&\textbf{5.6E-09}	&\textbf{5.2E-09}\\
&sd	&\textbf{7.5E-06}	&\textbf{2.9E-09}	&\textbf{2.8E-09}	&\textbf{3.1E-09}\\
		\end{tabular}
	\end{center}
\end{table}

\begin{algorithm}
	\caption{CPSO}
	\label{pseudoCPSO}
	\begin{algorithmic}[1]
		\State Initialize a population of particles with spatially optimized positions in \textit{D} dimensional of space.
		\State Require diffusion, directed motion/nucleation stepwise velocities.
		\State Require number of diffusion iterations ($DiffuseIter$).
		\State Require $MaxIter$.
		\State Require $ChaosMaxCount$ \Comment{Max iteration count before chaos factor is = 1}
		\State Require $ChaosMaxValue$ \Comment{Max influence ChaosFactor has on velocity}
		\While{Not Converged}
		\State iter++
		
		\For{Each particle $i$}
		
		\State $\overrightarrow{X}_i = \overrightarrow{X}_i + \overrightarrow{V}^{Total}_i$
		\State $Objective = f(\overrightarrow{X}_i)$
		
		\If{$iter \geq DiffuseIter$}
		\State $\overrightarrow{V}^{Total}_i = \overrightarrow{V}^{Directed}_i  ChaosFactor$
		\If{$Objective \geq GlobalBest$}
		\State $\overrightarrow{X}_i {{-}{=}} runif(0,1) \overrightarrow{V}^{Directed}_i \cdot e_i$
		\Else
		\State $\overrightarrow{X}_i {{+}{=}} runif(0,1) \overrightarrow{V}^{Directed}_i \cdot e_i$
		\EndIf
		
		\If{$(\overrightarrow{X}_i- \overrightarrow{g}_i) \geq 2e_i$}
		
		\State $\overrightarrow{V}^{Total}_i = runif(0,1) \overrightarrow{V}^{Total}_i \cdot e_i$
		\Else
		\State $\overrightarrow{V}^{Total}_i {{+}{=}} runif(0,1) \overrightarrow{V}^{Total}_i \cdot e_i$
		\EndIf
		
		\Else
		\State $\overrightarrow{V}^{Total}_i = \overrightarrow{V}^{Diffuse}_i  ChaosFactor$
		\State $GlobalBest = Objective$
		\EndIf

		\EndFor
		\State $ChaosFactor = ChaosMaxValue- ( (ChaosMaxValue-1) / (1 + exp(- ((Iter - (ChaosMaxCount*2/pi))/ (ChaosMaxCount/(2pi)) )   )))$
		\EndWhile
	\end{algorithmic}
\end{algorithm}

\end{document}